\documentclass[12pt]{spieman}  
\usepackage{amsmath,amsfonts,amssymb}
\usepackage{graphicx}
\usepackage{setspace}
\usepackage{tocloft}
\usepackage{multirow}
\usepackage{float}
\usepackage{booktabs}

\title{StegaINR4MIH: steganography by implicit neural representation for multi-image hiding}

\author[a,b]{Weina Dong}
\author[a,b,*]{Jia Liu}
\author[a,b]{Lifeng Chen}
\author[a,b]{Wenquan Sun}
\author[a,b]{Xiaozhong Pan}
\author[a,b]{Yan Ke}
\affil[a]{College of Cryptography Engineering, Engineering University of PAP, Xi '’an Shaanxi710086, China}
\affil[b]{Key Laboratory of Network and Information Security of PAP (Engineering University of PAP), Xi '’an Shaanxi 710086, China}

\cftpagenumbersoff{figure}
\cftpagenumbersoff{table} 
\begin{document} 
\maketitle

\begin{abstract}
Multi-image hiding, which embeds multiple secret images into a cover image and is able to recover these images with high quality, has gradually become a research hotspot in the field of image steganography. However, due to the need to embed a large amount of data in a limited cover image space, issues such as contour shadowing or color distortion often arise, posing significant challenges for multi-image hiding. In this paper, we propose StegaINR4MIH, a novel implicit neural representation steganography framework that enables the hiding of multiple images within a single implicit representation function. In contrast to traditional methods that use multiple encoders to achieve multi-image embedding, our approach leverages the redundancy of implicit representation function parameters and employs magnitude-based weight selection and secret weight substitution on pre-trained cover image functions to effectively hide and independently extract multiple secret images. We conduct experiments on images with a resolution of  from three different datasets: CelebA-HQ, COCO, and DIV2K. When hiding two secret images, the PSNR values of both the secret images and the stego images exceed 42. When hiding five secret images, the PSNR values of both the secret images and the stego images exceed 39. Extensive experiments demonstrate the superior performance of the proposed method in terms of visual quality and undetectability. 
\end{abstract}

\keywords{data hiding, steganography, implicit neural representation, multimedia security}

{\noindent \footnotesize\textbf{*}Second Author,  \linkable{liujia1022@gmail.com} }

\begin{spacing}{2}   

\section{Introduction}

Steganography is a technique used to hide secret information within multimedia data, such as image\cite{1}, videos\cite{2}, and text\cite{3}, and is widely applied in the fields of secret communication and copyright protection. With the rapid development of deep learning technology, steganography methods based on deep neural networks (DNN) \cite{4,5,6,7}have become a research focus. Traditional DNN steganography methods typically consist of a secret encoder and a secret decoder, where the sender uses the encoder to embed secret information into a cover image, and the receiver uses the decoder to extract the secret information from the stego image. Deep steganography views deep neural networks as a data processing tool, leveraging their powerful encoding and decoding capabilities to achieve higher levels of concealment and image quality. However, despite the significant progress made by deep neural network-based steganography, there are still some shortcomings.

Firstly, deep steganography methods often directly handle discrete image data, leading to rapid model expansion with increasing image resolution. This not only increases the training time and computational resource requirements but also poses challenges for processing high-resolution images. Secondly, these methods usually can only handle specific types of media data, lacking cross-type media generality. For instance, separate models need to be trained for images, audio, or videos, greatly increasing the complexity of the models. Lastly, deep steganography methods typically require an additional decoder to extract secret information, with a typical encoding-decoding network occupying over 100MB of storage space, which poses security risks for both message senders and recipients.

Some researchers have proposed steganographic techniques aimed at deep network model data to address the issue of covert transmission by decoders\cite{8,9,10,11,12,13}. These methods essentially treat model data as explicit carrier data consistent with multimedia data, typically modifying the model's weights and architecture using traditional carrier modification strategies to hide secret messages or functionalities. However, these methods still cannot avoid the problem of high communication overhead. Unlike traditional uses of deep neural networks as tools for processing a specific type of data object, implicit neural representation (INR)\cite{14} technology employs neural networks to represent a particular multimedia object. Various types of multimedia data can be transformed into implicit neural network representations. Unlike conventional discrete representation methods, INR can achieve high-precision image representation with a limited number of parameters, making it an emerging direction in steganographic research. Specifically, in INR, data is represented as continuous functions, meaning that the same neural network can handle data of different resolutions and embed high-resolution secret information without significantly increasing computational resources. Additionally, INR provides a more flexible way of data representation, allowing a single neural network to process different types of data, such as images, audio, and videos, enabling multimodal steganography. Finally, INR steganography does not require training an additional decoder; by converting secret data and cover data into a unified function representation, efficient and secure data transmission can be achieved.

Based on the existing research on steganography utilizing implicit neural representations , it is evident that the implementation of steganography on implicit representations of multimedia is still in its early stages. In 2023, Han et al.\cite{15} first introduced a deep cross-modal steganography framework using INR. Liu et al.\cite{16} introduced the concept of functional steganography based on implicit neural representations; however, images represented by implicit functions exhibited a high degree of distortion. Yang et al.\cite{17} proposed the INRSteg scheme, which can effectively hide multiple pieces of data without altering the original INR. In 2024, Dong et al.\cite{18} proposed a neuron pruning-based steganography scheme using implicit neural representations building on Liu's work; however, this scheme could only hide a single secret image.

To achieve high-quality multi-image hiding, we propose a novel Implicit Neural Representation Steganography framework. This approach is fundamentally different from existing methods12, \cite{12,19,20,21,22}, which typically utilize multiple encoders to achieve multi-image embedding. Instead, our method builds upon image implicit representation and fully leverages the redundancy of implicit representation function parameters to achieve large-capacity, high invisibility multi-image hiding. Specifically, we first represent the cover image implicitly, obtaining the cover function that represents the cover image. We then employ a magnitude-based weight selection method to select the importance weights for the cover function. Importance weights are identified using a marking mask, and important weight parameters in the cover function are replaced with secret weights generated from different random number seeds, while the remaining weight parameters are shared among multiple image representations. Experimental results demonstrate that this method effectively hides and independently extracts multiple images while maintaining high-quality image representation. Furthermore, security of secret information transmission is enhanced through different key-trigger mechanisms.

\begin{figure}[H]
    \centering
    \includegraphics[width=\linewidth]{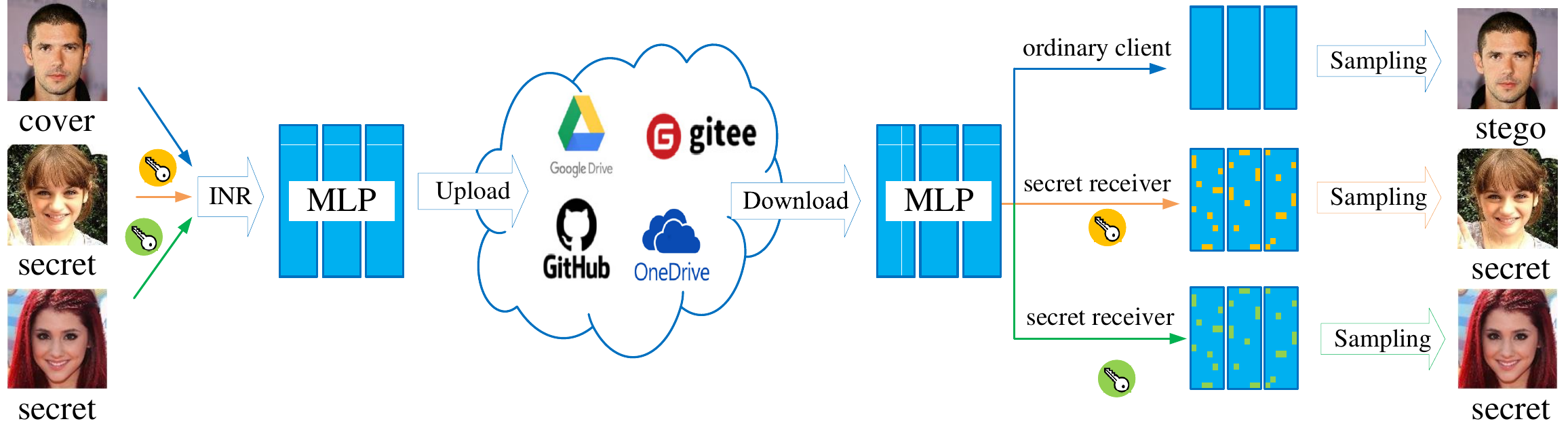}
    \caption{Illustration of the application scenario of the proposed StegaINR4MIH.}
    \label{fig1}
\end{figure}

In this framework, as shown in Fig. \ref{fig1}, the sender can publish the cover image and the secret image to be transmitted in the form of an implicit representation network in a public model repository. Ordinary users can download this implicit representation network to sample stego images, while secret users with keys can recover the secret network from the implicit network to sample secret images.

The main contributions of this paper are as follows:

1. We propose the StegaINR4MIH framework, which enables the effective hiding and independent extraction of multiple images through a single neural network function.

2. We design an innovative magnitude-based weight selection and secret weight replacement strategy that effectively prevents unauthorized recovery of the steganographic function without using side information. Additionally, pretraining the cover image function to initialize the function weights for the stego image significantly improves the representation quality of the stego image.

3. We apply implicit neural representations to steganography, effectively addressing issues such as the high training costs of deep steganography, heavy decoding transmission burdens, and limitations imposed by image resolution.

\section{Related Work}

\subsection{DNN-based Steganography}

The steganography schemes based on deep neural networks (DNNs) mainly adopt an encoder-decoder structure to achieve data embedding and extraction. Buluja et al.\cite{4} first proposed the use of an encoder-decoder network to hide and recover data, enabling the hiding of a secret image into a carrier image and the recovery of the secret image from the stego image through the decoder network. Weng et al.\cite{23} further applied this technology to video steganography through temporal residual modeling. Zhang et al.\cite{24} introduced a discriminator on top of the encoder-decoder network and utilized adversarial training to achieve a peak embedding rate of 4.4 bits per pixel (bpp). Zhu et al.\cite{7} added a noise layer to the encoder-decoder network, enabling high-quality recovery of the secret image from the stego image even under various noise attacks. Tancik et al.\cite{25} integrated the process of printing and recapturing images into the encoder-decoder, enhancing the performance of the secret decoder which demonstrates robustness against attacks arising from printing and recapturing processes. Additionally, Jing et al.\cite{5} pioneered the concept of combining reversible neural networks (INN) with steganography, modeling the recovery of secret messages as the inverse operation of message hiding. Subsequently, more researchers have been devoted to the design of reversible steganography networks\cite{6,21,22} leveraging the same encoding-decoding network for information hiding and extraction simultaneously.

Compared to the low-capacity information hiding of a single secret image, image hiding is more challenging due to its requirement for high capacity. In order to achieve larger hiding capacity, Baluja et al.\cite{19} first attempted to concatenate two secret images and hide them into a cover image using an autoencoder-based network. Building on this, Das et al.\cite{20} proposed a similar MISDNN structure based on autoencoder for multiple image hiding. Lu et al.\cite{21} introduced a reversible network ISN that utilizes the same encoding-decoding network for information hiding and extraction concurrently. Guan et al.\cite{22} further developed the DeepMIH framework to enable hiding of multiple images.

Although the aforementioned deep steganography schemes possess significant advantages over traditional methods, such as higher secrecy, larger capacity, and robustness, they still face several challenges. Firstly, these networks are typically trained on discrete grid data, leading to the size of the encoders and decoders being directly associated with the resolution of the underlying grid data, necessitating a large amount of sample data and computational resources for model training. Secondly, deep steganography schemes often require transmitting the decoder secretly to the receiver, increasing communication overhead and the risk of exposing steganographic activities.

\subsection{DNN Model Steganography}

To address the issue of covert transmission of decoders, researchers have begun focusing on the problem of neural network model covert communications and have proposed DNN model steganography, with the aim of using neural network model data as carriers for information hiding. Yang et al. \cite{8} embed secret data into the convolutional layers of a given neural network (cover network) during network training. In a network containing the secret data (stego network), matrix multiplication is used to encode the parameters of the convolutional layers for data extraction. Chen et al.\cite{9} utilized DNN to establish a probability density model of masked images and hid a secret image at a specific location in the learned distribution. They used SingGAN to learn the patch distribution of a cover image and hid the secret image by fitting a deterministic mapping from a fixed noise mapping (generated by the embedding key). Yang et al.\cite{10} proposed a neural network multi-source data iding scheme where multiple senders can simultaneously send different secret data to a receiver using the same neural network. They achieved data embedding during the training process of the neural network, replacing post-training modifications to the neural network. These methods view the neural network model as an explicit data similar to videos, audios, and images. Li et al.\cite{11,12,13} introduced the concept of covert DNN models, hiding secret DNN models into covert DNN models performing ordinary machine learning tasks, designed to integrate with tasks such as image classification, segmentation, or denoising to ensure the security of the message decoder. However, the above methods still cannot avoid the problem of high communication overhead. Furthermore, in the schemes 11 and 12, the neuron positions of the secret decoding network in the steganographic model are recorded as side information for model recovery, making the recovery process inconvenient for practical applications.

\subsection{Implicit Neural Representation Steganography}

Implicit neural representation (INR) steganography is not the conventional neural network model steganography, but a technique that converts multimedia data from explicit representations to parameterized function representations. Essentially, it involves performing steganographic operations on implicit representation models (functions). INR offers advantages such as independence from spatial resolution, strong representation capability, strong generalization ability, and ease of learning, and has been widely applied to various two-dimensional and three-dimensional visual tasks, including images\cite{26}, videos\cite{27}, 3D shapes\cite{28}, 3D scenes\cite{29}, and 3D structure appearances\cite{30}, providing new methods for steganography. Neural Radiance Fields (NeRF)\cite{30} uses neural networks to implicitly simulate three-dimensional scenes, attracting extensive research attention. In 2022, StegaNeRF\cite{31} conducted the first investigation into information hiding in NeRF. They hid natural images into 3D scene representations by retraining NeRF parameters, simultaneously training a decoder capable of accurately extracting hidden information from 2D images rendered by NeRF. In 2023, CopyRNeRF\cite{32} focused on copyright protection for NeRF. They protected NeRF models' copyrights by replacing the original color representation in NeRF with a watermark color representation. They used a decoder to recover binary secret information from rendered images while maintaining high rendering quality and allowing watermark extraction. In 2024, a series of copyright and steganographic schemes for NeRF emerged\cite{33,34,35}. These methods all utilize neural radiance fields as image generators and, similar to DNN steganography, require training separate decoders for effective message extraction.

Han et al.\cite{15} proposed a deep cross-modal steganography framework using INR, capable of handling various modes and resolutions of data, effectively overcoming the limitation of DNN steganography focusing only on specific data types. However, this method embeds implicit model data into the explicit representation carrier of images, still employing traditional multimedia steganography methods. Liu et al.\cite{16} introduced the concept of function steganography based on implicit neural representation, embedding the implicit neural representation function of secret information into the carrier data's implicit neural representation function. This allows direct extraction of secret messages from the stego function using a shared key between sender and receiver, without the need for a decoder, ensuring higher concealment. Yang et al.\cite{17} concatenated multiple message implicit representation models and carrier implicit representation models onto a single large network based on message implicit representation and scrambled them. However, the scrambled steganographic network can only output noisy data, resulting in lower concealment for steganography. Building upon Liu's\cite{16} work, Dong et al.\cite{18} proposed a neuron pruning-based implicit neural representation steganographic scheme, training an INR function simultaneously representing stego images and secret images, effectively enhancing the quality of secret and stego images represented by neural network parameters. 

In this paper, we convert both secret data and steganographic data into a unified data format, namely functions, and propose a steganographic framework that enables the hiding of multiple images through a single neural network function. Compared to deep steganography schemes based on encoder-decoder networks, our method leverages the unique advantages of implicit representations to effectively address issues such as high training costs, heavy decoding transmission burdens, and limitations imposed by image resolution. In contrast to model-based steganography schemes, our approach does not require training an additional decoder, thereby avoiding high communication overhead. Additionally, while schemes 11 and 12 rely on auxiliary information for recovering the secret network, our method devises an innovative amplitude-based weight selection and secret weight replacement strategy, which effectively prevents unauthorized recovery of the steganographic function without the use of auxiliary information. Compared to existing implicit neural representation steganography schemes, our method achieves dual implicit representation of secret data and steganographic data, enabling a greater capacity for multi-image implicit steganography, while also providing higher image representation quality and a more covert key-trigger mechanism.

\section{The Proposed Method}

\subsection{framework}

\begin{figure}[H]
    \centering
    \includegraphics[width=\linewidth]{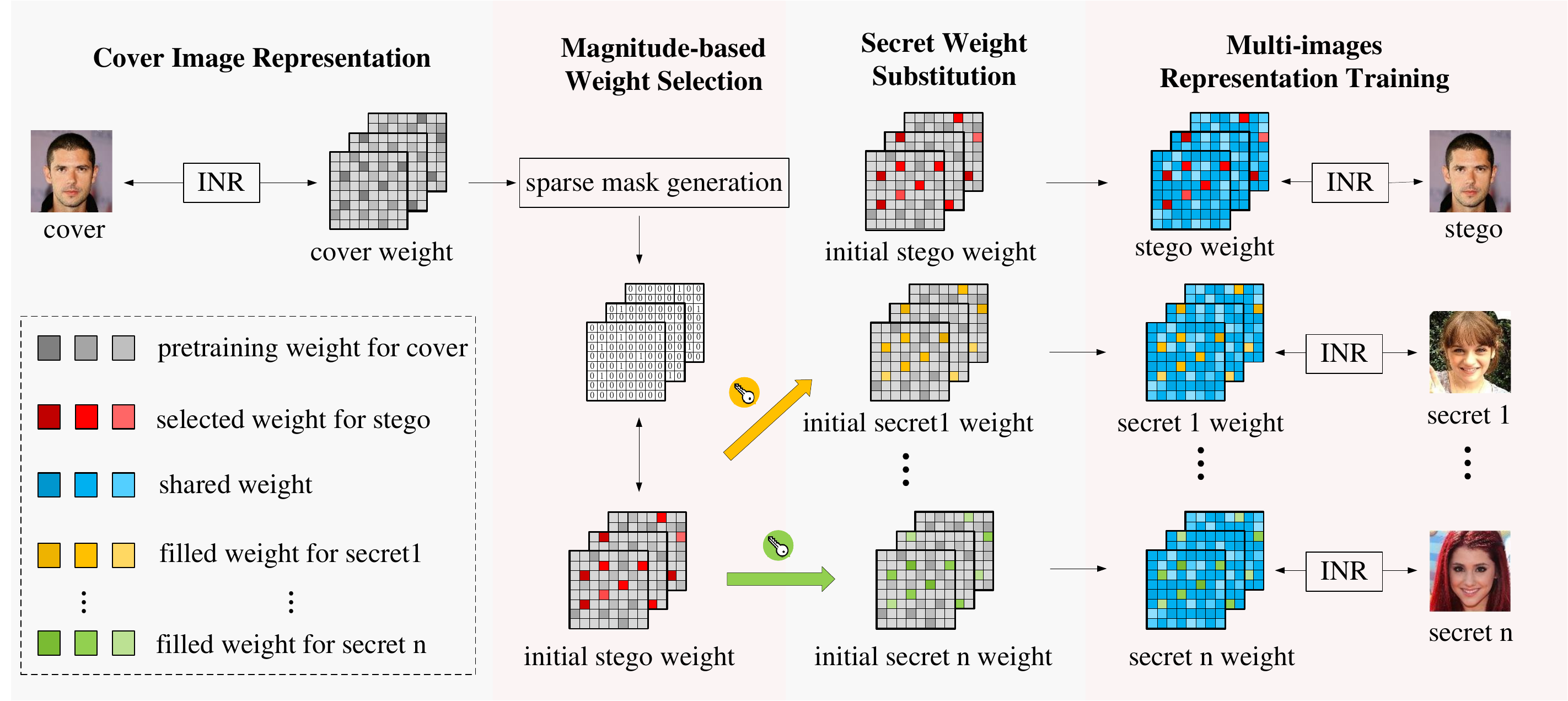}
    \caption{An overview of our proposed method.}
    \label{fig2}
\end{figure}

Our research focuses on achieving multi-image steganography using Implicit Neural Representation (INR), converting all data representations into implicit neural representations (INR) as in the methods of Siren Sitzmann et al.\cite{14}, and implementing steganography and representation of multiple images through a single neural network function. Fig. \ref{fig2} illustrates the overall framework of our approach, which is divided into four sub-stages. \textbf{Step 1: Cover image representation.} The cover image is first converted into its corresponding implicit neural representation. \textbf{Step 2: Magnitude-based weight selection.} A magnitude-based weight selection method is employed to identify important weights (shown in red in the Fig. \ref{fig2}) of the pre-trained cover function through sparse mask generation. \textbf{Step 3: Secret weight substitution.} Important weights selected in Step 2 are replaced with weights initialized from different random seeds (shown in orange and green in the Fig. \ref{fig2}). \textbf{Step 4: Multi-image representation training.} In this stage, the stego function is a reconstructed form of the cover function. By fixing the important weights of the stego function and the secret weights of multiple secret functions, the shared blue weights are jointly trained. This enables representation of multiple images through a single architecture. The stego function, while representing stego images, can be triggered to represent different secret images using different keys.

\subsection{Cover image representation}

\begin{figure}[H]
    \centering
    \includegraphics[width=0.3\linewidth]{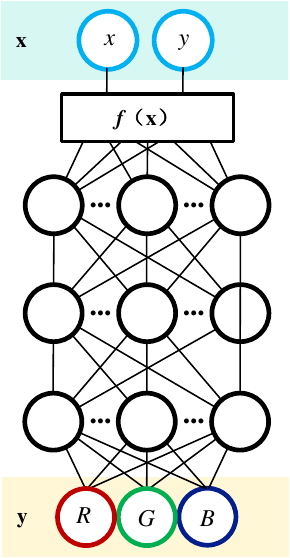}
    \caption{The implicit neural representation of image signals.}
    \label{fig3}
\end{figure}

Implicit neural representation is a novel method for parameterizing various signals, including images, audio, video, and 3D models. It parameterizes these signals as a continuous function, mapping the domain of the signal to the values of properties at those coordinates, which is also referred to as coordinate-based representation. This representation no longer relies on traditional data formats, such as networks or voxels, for storing and processing information, but rather utilizes the neural network itself as the representation of the data. For images, each pixel within a plane of the image has coordinates (x,y), with each coordinate corresponding to the RGB value of that pixel. By learning the mapping relationship between the coordinates and the RGB values through a neural network function, we obtain the implicit neural representation of the image, as shown in Figure \ref{fig3}.

In the first stage, we represent the cover data by fitting INR to the cover image. The structure of the INR is a multi-layer perceptron (MLP) network with activation function $\sigma(\cdot)$, consisting of n hidden layers of size D, where the output can be represented as equation (1):
\begin{equation}
    \mathbf{y}=\mathbf{W}^{(n)}(g_{n-1}\circ\cdots\circ g_{1}\circ g_{0})(\mathbf{x}_{0})+\mathbf{b}^{(n)},\\where \quad \mathbf{x}_{i+1}=g_{i}(\mathbf{x}_{i})=\sigma(\mathbf{W}^{(i)}\mathbf{x}_{i}+\mathbf{b}^{(i)})
\end{equation}
where $x_i$ represents the input to the $i^{th}$ hidden layer,$i\in\left\{0,1,2,\cdots,n\right\}, y\in\mathbb{R}^{0}$, corresponds to the output of $\mathbf{x}_{0}\in\mathbb{R}^{I}$, and $g_{i}:\mathbb{R}^{D}\to\mathbb{R}^{D}$. $\mathrm{W}^{(n)}$ denotes the weight matrix of the $n$-th layer and $g_{i}(\mathrm{x}_{i})$ represents the activation function, indicating that the input $x_0$ is successively transformed by the activation functions from layer 0 to layer $n$-1. We use SIREN (Sinusoidal Representations for Neural Networks) as the base structure of the INR network, which utilizes a sinusoidal activation function widely used in the field of INR research. The implicit neural representation of the cover image is learned by minimizing the following loss function:
\begin{equation}
    \mathcal{L}_{\mathrm{cov} er}=\min_{\theta}\sum_{(\mathrm{x},\mathrm{y})\in S}\left\|f_{\theta}\left(\mathrm{x}\right)-\mathrm{y}\right\|_{2}^{2}
\end{equation}
where $f_{\theta}$ represents the implicit neural representation of the cover image, and the set $S$ consists of 2D coordinates $\mathbf{x}\in\mathbb{R}^{2}$ and corresponding RGB values $\mathbf{y}\in\mathbb{R}^{3}$.

\subsection{Magnitude-based weight selection}

We select the weights of the pre-trained cover image based on their contributions to representing the cover image, in order to achieve higher quality stego image representation in subsequent joint training. Therefore, we propose a magnitude-based weight selection method to generate a sparse mask, given a fixed ratio S (0<S<1) and the total number of weights N in the cover function. The fixed ratio S refers to the proportion of weights that are fixed in the entire network. The sparse mask is generated through the following process:

(1) Sort the weight parameters of the pre-trained implicit representation of the cover image in descending order of their absolute values.

(2) Introduce a binary mask matrix M of the same size as the weight matrix W, and calculate the mask M as follows:
\begin{equation}
    \mathbf{M}_{i,j}=\begin{cases}1&\text{if}\mid W_{i,j}\mid>\mathbf{t}\\0&\text{otherwise}\end{cases}
\end{equation}
where $W_{i,j}$ represents the weight parameter in the weight matrix of the pre-trained cover function, and $M_{i,j}$ represents an element in the mask matrix. t is the threshold for the p-th largest weight in W. We denote the weights corresponding to elements in the mask matrix that are 1 as $W_{st}$, which are important weights for the stego function training. The weights corresponding to elements in the mask matrix that are 0 are denoted as $W_s$ and are shared with the embedding secret function.

  Therefore, the StegaINR4MIH framework can be represented as:
  \begin{equation}
      \mathcal{F}(\mathrm{W}_{st}\odot\mathrm{M}+\mathrm{W}_{s}\odot\overline{\mathrm{M}})\to x_{st}
  \end{equation}
  Where $\mathcal{F}$ represents the function for implicit neural representation of images. $\odot$ represents the element-wise product of the weight matrix W and the mask matrix M. $\overline{\mathbf{M}}$ represents the complement matrix of the mask matrix M. x$_{st}$ represents the stego image.

\subsection{Secret Weight Substitution}

In the secret weight replacement stage, the selected important weights $W_{st}$ are replaced by new, randomly initialized secret weights $W_{se}$. Each secret image corresponds to a set of secret weights, which are initialized using independent random seeds to ensure independence and confidentiality between different secret images.

The sender replaces the important weights $W_{st}$ with secret weights $W_{se}$ using the following formula:
\begin{equation}
    \mathcal{F}(\mathrm{W}_{se}\odot\mathrm{M}+\mathrm{W}_{s}\odot\overline{\mathrm{M}})\to x_{se}
\end{equation}
where $W_{se}$ is a set of random weights initialized as follows:
\begin{equation}
    \mathrm{W}_{se}=\mathrm{I}(N(\cdot),k_{se}), \mathrm{W}_{se}\sim N(0,\frac{2}{n_{in}+n_{out}})
\end{equation}
where $\mathrm{I}(\cdot)$ is the random initialization used for key seeding, weights initialized with the same random seed are identical. $k_{se}$ serves as the key to obtain the initialization of secret weights. We use the Xavier\cite{36} algorithm to initialize the secret weights $W_{se}$, which means that each element of $W_{se}$ is randomly sampled from a normal distribution with a mean of 0 and a variance of $\frac{2}{n_{in}+n_{out}}$.
Here, $n_{in}$ and $n_{out}$ represent the number of input and output neurons, respectively.

\subsection{Multi-images representation training}

To efficiently achieve steganography and representation of multiple images through a single neural network function, the training consists of two main losses: stego function loss and multi-secret function loss.

\textbf{Stego Function Loss:} This loss aims to make the stegp images represented by neural implicit function more realistic, with the loss function defined as:
\begin{equation}
    \mathcal{L}_{stego}=\min_{\varphi}\sum_{(\mathrm{x,y})\in S}\left\|f_{\varphi}\left(\mathrm{x}\right)-\mathrm{y}\right\|_{2}^{2}
\end{equation}
where $f_{\varphi}$ represents the implicit neural representation of the stego image, and the set S consists of 2D coordinates $\mathbf{x}\in\mathbb{R}^{2}$ and corresponding RGB values $\mathbf{y}\in\mathbb{R}^{3}$.

\textbf{Multi-Secret Function Loss:} This loss aims to train the stego function to hide multiple secret functions using the implicit representation, thus concealing multiple secret images. The loss function is given by:
\begin{equation}
    \begin{aligned}&\mathcal{L}_{sec.ret}=\operatorname*{min}_{\varphi_{1}}\sum_{(\mathbf{x},\mathbf{y})\in S_{1}}\left\|f_{\varphi_{1}}\left(\mathbf{x}_{1}\right)-\mathbf{y}_{1}\right\|_{2}^{2}+\operatorname*{min}_{\varphi_{2}}\sum_{(\mathbf{x},\mathbf{y})\in S_{2}}\left\|f_{\varphi_{2}}\left(\mathbf{x}_{2}\right)-\mathbf{y}_{2}\right\|_{2}^{2}\\&+\cdots+\min_{\varphi_{N}}\sum_{(\mathbf{x},\mathbf{y})\in S_{N}}\left\|f_{\varphi_{N}}\left(\mathbf{x}_{N}\right)-\mathbf{y}_{N}\right\|_{2}^{2}\end{aligned}
\end{equation}
where $f_{\varphi}(\cdot)$ represent the implicit neural representations of the corresponding secret images, and the set $S$ consists of 2D coordinates $\mathbf{x}\in\mathbb{R}^{2}$ and corresponding RGB values $\mathbf{y}\in\mathbb{R}^{3}$ for different secret images. $N$ is the number of secret images.

To balance the quality of representation for stego and secret images, optimization is performed using a specific joint loss function:
\begin{equation}
\mathcal{L}_{total}=\lambda_{st}\mathcal{L}_{stego}+\lambda_{se}\mathcal{L}_{secret}
\end{equation}
where $\lambda_{st}$ and $\lambda_{se}$ represent the hyperparameters of the stego and secret functions, respectively.

During training, only the blue weights in Figure \ref{fig2} (denoted as $W_s$) are updated, which are shared between the stego function and multiple secret functions. By fixing the important weights of the cover image $W_{st}$ and initializing and replacing them with secret weights $W_{se}$ at the important weight positions of $W_{st}$ using random seeds, updating only the shared portion of the weights $W_s$ among the stego and different secret functions optimizes StegaINR4MIH to achieve the functionality of representing multiple images through a single network architecture. Given a learning rate $\alpha$, the weights $W_{s}$ are updated using gradient descent.
\begin{equation}
    \mathrm{W}_{s}=\mathrm{W}_{s}-\alpha(\lambda_{st}\nabla_{\mathrm{W}_{s}}\mathcal{L}_{stego}+\lambda_{se}\nabla_{\mathrm{W}_{s}}\mathcal{L}_{secret})
\end{equation}

\subsection{Sparse Representation of Mask Matrix}
\begin{figure}[H]
    \centering
    \includegraphics[width=0.8\linewidth]{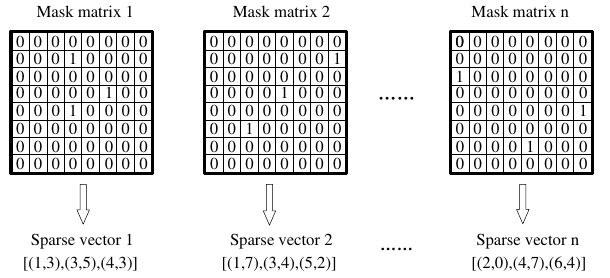}
    \caption{Converting mask matrices to sparse vectors.}
    \label{fig4}
\end{figure}

In this framework, in addition to pre-arranging random seeds between the sender and secret recipient, the sender also needs to transmit the mask matrix derived from the pre-trained cover function based on weight magnitudes to the receiver in a confidential manner. However, in practical applications, transmitting this edge information may lead to excessive key transmission volume, increasing transmission risks. Therefore, we adopt a sparse representation method for the mask matrix, whereby the mask matrix is transformed into a more compact form using compression techniques. This is because our method only requires fixing a small number of weight parameters with larger magnitudes, thus only the positions of these fixed weights, where the mask is 1, need to be transmitted, as shown in Fig. \ref{fig4}.

For each mask matrix, identify all positions with a value of 1, which can be represented using the row and column indices of the matrix, forming a sparse vector. Assuming n=8, if the size of the mask matrix to be transmitted before compression is 512 bits, after sparsifying, only 48 bits of information are needed, achieving a compression rate of up to 90.62\%. Therefore, by converting multi-layer 2D mask matrices into one-dimensional sparse vectors and retaining only the indices of positions where the mask value is 1, the amount of transmitted data can be significantly reduced, enhancing transmission security and efficiency.

\subsection{INR Image Recovery}
The stego function can be transmitted over a network's public channel, and regular users can only sample the stego images represented by the stego function without detecting the presence of secret images. Since we jointly train the stego function and multiple secret functions by replacing key weights without altering the parameters of the secret functions, secret recipients can seamlessly recover the secret functions using the key K after obtaining the stego function. The key K is composed of the sparse vector of the cover function and the random seed. For different recipients from the same sender, their sparse vectors are the same but the random seeds are different. Different random seeds initialize different weights, corresponding to different secret images. Initially, secret weights $W_{se}$ are initialized by the random seed, and then the weights $W_{st}$ are replaced by the secret weights $W_{se}$ using the formula below to obtain the parameters of the corresponding secret functions.
\begin{equation}
    \mathcal{F}(\mathrm{W}_{se}\odot\mathrm{M}+\mathrm{W}_{s}^{\prime}\odot\overline{\mathrm{M}})\to x_{se}
\end{equation}
Where $\mathrm{W}_{s}^{\prime}$ (the blue weights in Figure \ref{fig2}) refers to the updated weights of $W_s$(the gray weights in Figure \ref{fig2}).

The process of sampling secret images from the secret function is equivalent to sampling explicit data from the implicit representation function, as shown in the formula below. The recipient only needs to input a set of coordinates $\left\{\left(x_{i},y_{i}\right)_{i=1}^{n}\right\}$ to obtain the corresponding RGB pixel values $\left\{p_{i}\right\}_{i=1}^{n}$.
\begin{equation}
    \left\{p_{i}\right\}_{i=1}^{n}=f\left\{\left(x_{i},y_{i}\right)_{i=1}^{n}\right\}
\end{equation}
Here, $x_i$ and $y_i$ represent the x and y coordinates of the $i^{th}$ pixel, n represents the total number of pixels, and $i$ denotes the pixel index.

\begin{figure}[H]
    \centering
    \includegraphics[width=\linewidth]{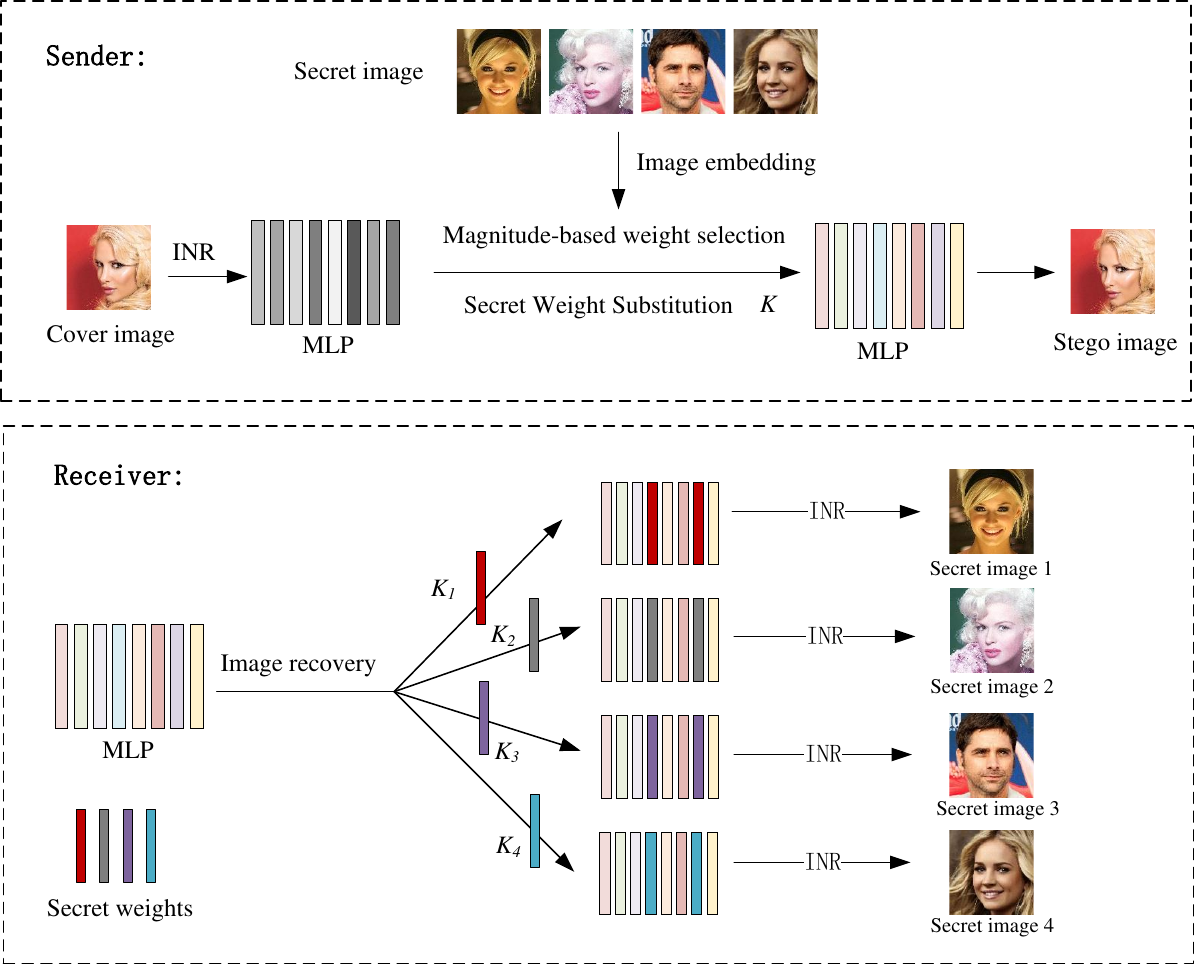}
    \caption{The entire process of the proposed scheme.}
    \label{fig5}
\end{figure}

Figure \ref{fig5} illustrates the entire process of the proposed scheme. Within this framework, the sender can hide multiple secret images within a cover image. The stego images are transmitted over the public channel of the network in the form of a stego function. The secret receiver only needs to possess a pre-agreed key to recover the corresponding secret images from the stego function.

\section{Experimental and Results Analysis}

\subsection{Experimental Settings}

In this work, we use the PyTorch 1.7.0 framework on a server equipped with NVIDIA GeForce RTX2070, with CUDA version 11.6, and using the Python 3.8 coding language. Extensive experiments are conducted on images from the CelebAHQ\cite{37}, COCO\cite{38}, and div2k\cite{39} datasets, all resized to a resolution of 128$\times$128. We parameterize the cover images using a multi-layer perceptron with eight hidden layers, each containing 128 neurons. The input layer dimension is 2, representing pixel coordinates, and the output layer dimension is 3, representing RGB values. The parameterized training of cover images utilizes the ADAM optimizer with a learning rate of 1e-3, trained for approximately 20,000 epochs, and the best-performing model is selected for joint training of multiple image implicit representations. During the joint training process, the weight selection ratio S based on magnitude is set to 0.05, and the weight hyperparameters for the stego function and secret function are set to 1/N+1, where N is the number of secret images. The training process involves using stochastic gradient descent with a learning rate of 1e-3 for approximately 50,000 epochs.

\subsection{Evaluation Metrics}

To evaluate the distortion between secret data and sampled secret data, as well as stego data and sampled stego data, we visualize the implicitly represented data and use four evaluation metrics: Peak Signal to Noise Ratio (PSNR)\cite{40}, Structural Similarity (SSIM), Root Mean Square Error (RMSE), and Mean Absolute Error (MAE).

(1) Peak Signal to Noise Ratio (PSNR)

PSNR is a commonly used method for evaluating image reconstruction quality. Given two images X and Y of size W$\times$H, it is defined using the Mean Square Error (MSE) as follows:
\begin{equation}
    \begin{aligned}&MSE=\frac{1}{W\times H}\sum_{i=1}^{W}\sum_{j=1}^{H}\biggl[X_{i,j}-Y_{i,j}\biggr]^{2}\\&PSNR=10\times\log_{10}\frac{MAX^{2}}{MSE}\end{aligned}
\end{equation}
where $X_{i,j}$ and $Y_{i,j}$ represent pixel values at position (i, j) in images X and Y, respectively, and MAX is the maximum pixel value. A higher PSNR value indicates lower distortion.

(2) Structural Similarity (SSIM)

SSIM is a metric for evaluating image quality that measures similarity in terms of brightness, contrast, and structure. It is computed as:
\begin{equation}
    \begin{aligned}&l(x,y)=\frac{2\mu_{x}\mu_{y}+C_{1}}{\mu_{x}^{2}+\mu_{y}^{2}+C_{1}}\\&c(x,y)=\frac{2\sigma_{x}\sigma_{y}+C_{2}}{\sigma_{x}^{2}+\sigma_{y}^{2}+C_{2}}\\&s(x,y)=\frac{\sigma_{xy}+C_{3}}{\sigma_{x}\sigma_{y}+C_{3}}\end{aligned}
\end{equation}
where $\mu_x$, $\sigma_x$ represent the mean and variance of image X, $\mu_y$, $\sigma_y$ represent the mean and variance of image Y, and $\sigma_{xy}$ is the covariance between X and Y. Constants C$_1$, C$_2$, and C$_3$ are typically set as specified. 
\begin{equation}
    SSIM(X,Y)=l(x,y)\cdot c(x,y)\cdot s(x,y)
\end{equation}
SSIM values range from 0 to 1, with higher values indicating lower distortion.

(3) Root Mean Square Error (RMSE)

RMSE measures the sample standard deviation of the differences (residuals) between predicted and true values, akin to the L2 norm. It is calculated as:
\begin{equation}
    RMSE=\sqrt{MSE}
\end{equation}

(4) Mean Absolute Error (MAE)

MAE represents the average of the absolute errors between predicted and true values, akin to the L1 norm. Unlike RMSE, MAE focuses more on minor differences in images and is calculated as:
\begin{equation}
    MAE=\frac{1}{W\times H}\sum_{i=1}^{W}\sum_{j=1}^{H}\left|X_{i,j}-Y_{i,j}\right|
\end{equation}

Integrating these four evaluation metrics, the higher the PSNR and SSIM values, and the lower the RMSE and MAE values, the lower the distortion in the images. Furthermore, we employ two steganalysis tools, StegExpose\cite{41} and SiaStegNet\cite{42}, to evaluate the undetectability of stego images represented by the stego function. We also employ three strategies to detect the steganalysis of the stego function, aimed at identifying the presence of the secret function within the stego function.

\subsection{Visual Quality}

\begin{figure}[H]
    \centering
    \includegraphics[width=0.8\linewidth]{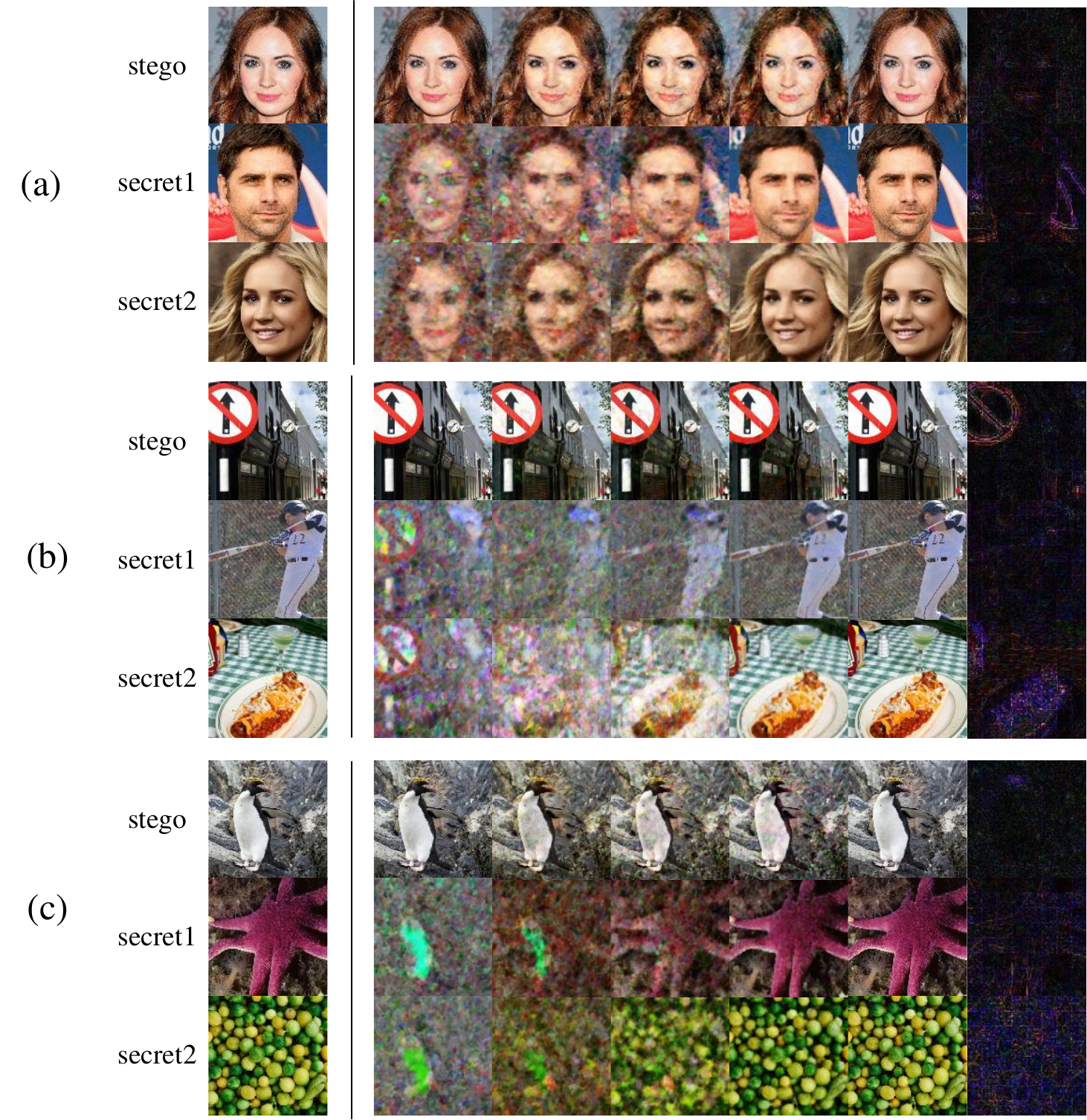}
    \caption{Visualisation of stego and secret images. (a) CelebAHQ  (b) COCO  (c) div2k}
    \label{fig6}
\end{figure}

Figure \ref{fig6} presents the experimental results of hiding two secret images on CelebAHQ, COCO, and div2k datasets from top to bottom. In order to demonstrate the optimization process of stego and secret image generation more clearly after secret weighting replacement, different qualities of generated images at various optimization iterations are displayed in Figure \ref{fig6}. The first column shows real images sampled from the dataset, while the second column displays sampled images after secret weighting replacement. The stego image is obtained by sampling from a pretrained cover function, whereas the secret image is sampled from the secret function with weights initialized using different random number seeds. Columns three, four, and five present images sampled after training for 5, 10, and 50 epochs, respectively, with column six showing images sampled at the end of training from the stego and secret functions. The last column shows the magnified differences between the sampled and real images.
From Fig. \ref{fig6}, it can be observed that after 10 epochs of optimization, the approximate outlines of the stego and secret images can be obtained. After 50 epochs, clearer images can be achieved, demonstrating that our proposed implicit neural representation method can quickly establish convergence relationships within a short period of time, enabling multi-image hiding. As the number of optimization iterations increases, the quality of the stego images initially decreases significantly before gradually improving. This is because only 5\% of important weights are retained in the joint training, requiring retraining of the remaining weights. Consequently, the remaining weights undergo a process of transitioning from initially fitting the cover image to simultaneously fitting the stego and secret images, resulting in a quality progression from poor to better and eventually reaching a balanced state with the secret image. The quality of the secret images significantly improves with increasing optimization iterations, with almost no visual differences between the original and sampled images.

We evaluate the image quality from three different datasets, with the experimental results shown in Table \ref{tab1}. It can be observed that due to the pretraining of cover images and retention of 5\% important weights, the image quality is significantly higher than that of the secret images, with PSNR performance above 45 on all three datasets. The PSNR of secret images is above 42 on all datasets, with CelebAHQ dataset outperforming the other two, possibly due to its simpler and more easily learnable texture structure leading to better performance.

\begin{table}[H]
    \centering
    \scriptsize
    \caption{Image Quality Under Different Datasets.}
    \begin{tabular}{l|llll|llll|llll}
\hline 
& \multicolumn{4}{|c|}{ Cele } & \multicolumn{4}{c|}{ COCO } & \multicolumn{4}{c}{ Div2k } \\
\cline { 2 - 13 } & PSNR$\uparrow$  & SSIM$\uparrow$  & RMSE$\downarrow$  & MAE$\downarrow$  & PSNR$\uparrow$  & SSIM$\uparrow$  & RMSE$\downarrow$  & MAE$\downarrow$  & PSNR$\uparrow$  & SSIM$\uparrow$  & RMSE$\downarrow$  & MAE$\downarrow$  \\
\hline stego & 46.86 & 0.9990 & 1.27 & 1.01 & 45.98 & 0.9987 & 1.19 & 1.22 & 45.62 & 0.9987 & 1.35 & 1.09 \\
Secret1 & 44.55 & 0.9969 & 1.51 & 1.14 & 45.53 & 0.9981 & 1.34 & 0.98 & 44.33 & 0.9983 & 1.54 & 1.16 \\
Secret2 & 45.42 & 0.9985 & 1.36 & 1.05 & 42.62 & 0.9980 & 1.88 & 1.37 & 42.39 & 0.9973 & 1.93 & 1.46 \\
\hline
\end{tabular}  
    \label{tab1}
\end{table}

\subsection{Multi-Image Steganography}

We conduct experiments on hiding more than two images using the same Implicit Neural Representation (INR) structure for stego data. As shown in Fig. \ref{fig7}, we perform experiments on hiding three, four, and five secret images using the CelebA dataset. The first column displays real images sampled from the dataset, the second column shows sampled images after secret weight replacement, the third column shows images sampled after 10 training epochs, the fourth column displays images sampled at the end of training from the stego and secret functions, and the last column presents magnified differences between the sampled images and the real images. From Fig. \ref{fig7}, it can be observed that even with an increased number of secret images, our approach maintains good visual quality in representing stego and secret images.

\begin{figure}[H]
    \centering
    \includegraphics[width=0.8\linewidth]{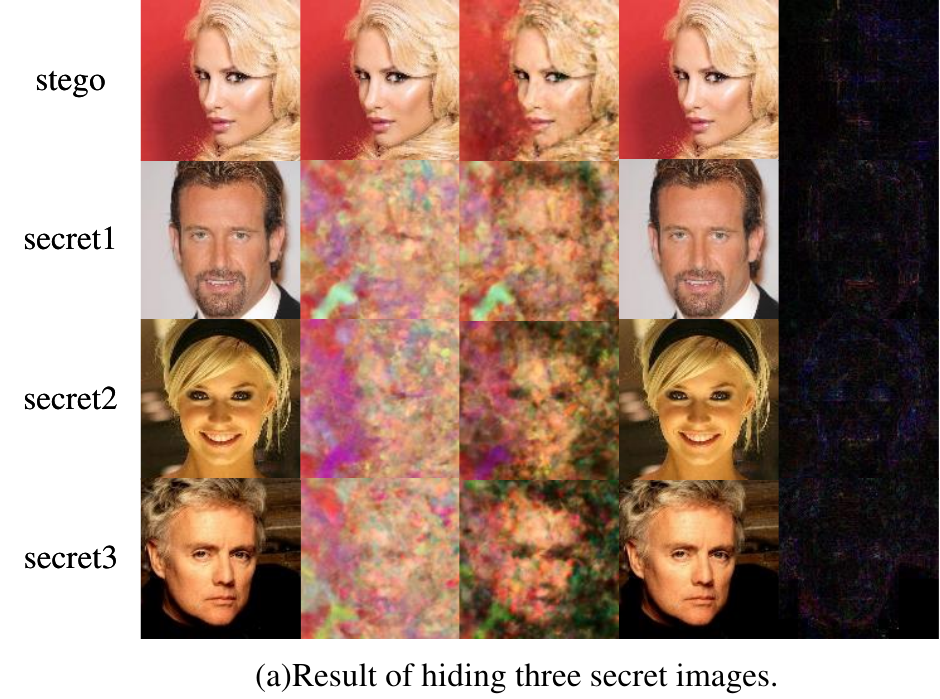}
    \includegraphics[width=0.8\linewidth]{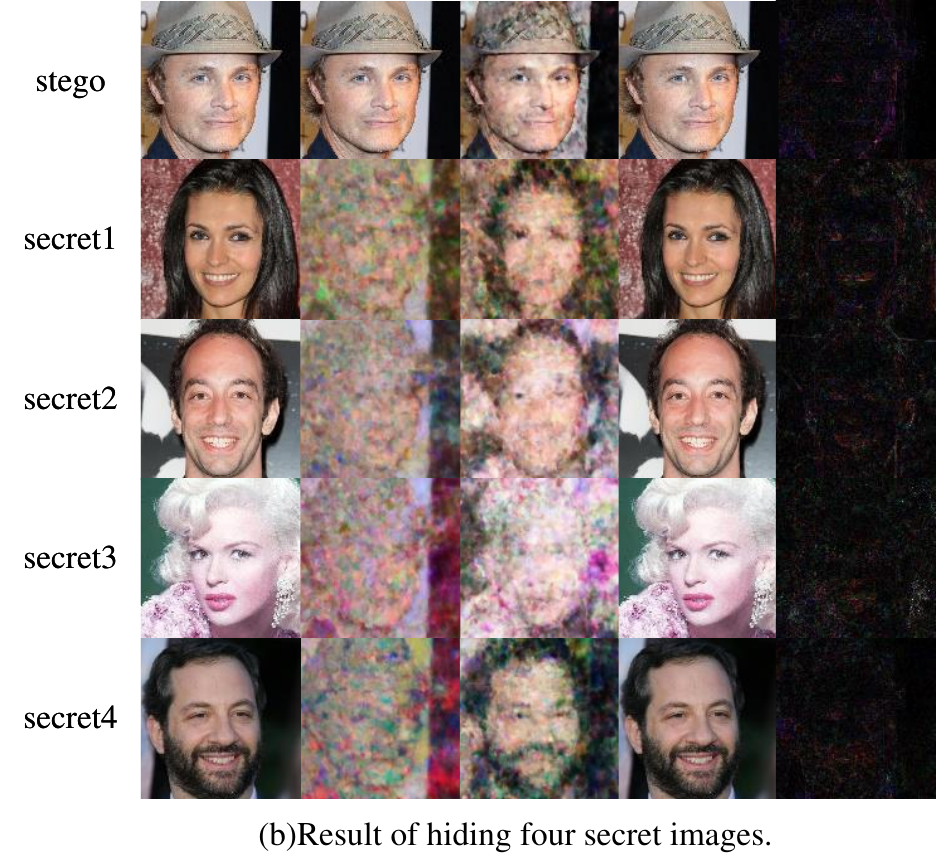}
\end{figure}

\begin{figure}[H]
    \centering
    \includegraphics[width=0.8\linewidth]{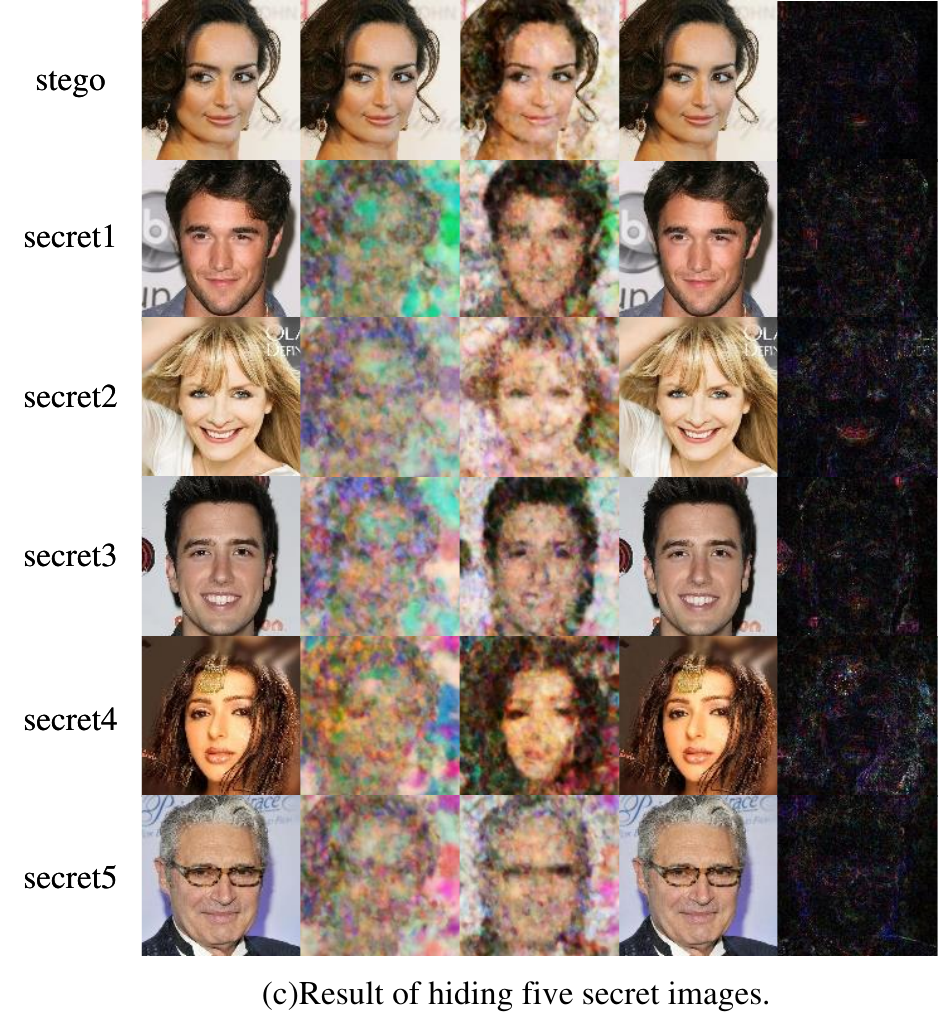}
    \caption{Visualisation of hiding multiple secret images.}
    \label{fig7}
\end{figure}

The quantitative evaluation results are presented in Table \ref{tab2}. With an increase in the number of hidden secret images, the quality of both stego and secret images slightly decreases. While our approach does not restrict the number of hidden images, increasing the amount of secret data leads to significantly higher training time and memory usage. Training complexity also increases as the number of hidden secret images grows. Therefore, it is advisable to determine the number of secret images based on practical needs and adjust the network structure size to accommodate memory capacity.

\begin{table}[H]
    \centering
    \caption{Image quality for hidden multiple images($>$ 2).}
    \begin{tabular}{c|c|cccc}
\hline & & \text { PSNR }$\uparrow$ & \text { SSIM }$\uparrow$ & \text { RMSE }$\downarrow$ & \text { MAE }$\downarrow$ \\
\hline 3 & \text { stego } & 47.77 & 0.9974 & 1.45 & 1.04 \\
\text { images } & \text { secret 1,2,3 } & 43.89 & 0.9926 & 1.63 & 1.21 \\
\hline 4 & \text { Stego } & 46.39 & 0.9972 & 1.22 & 0.81 \\
\text { images } & \text { secret 1,2,3,4 } & 42.09 & 0.9932 & 2.00 & 1.42 \\
\hline 5 & \text { stego } & 44.68 & 0.9946 & 1.48 & 1.04 \\
\text { images } & \text { secret1,2,3,4,5 } & 39.75 & 0.9904 & 2.66 & 1.76 \\
\hline
    \end{tabular}
    \label{tab2}
\end{table}

\subsection{Super-Resolution Sampling}

\begin{figure}[H]
    \centering
    \includegraphics[width=\linewidth]{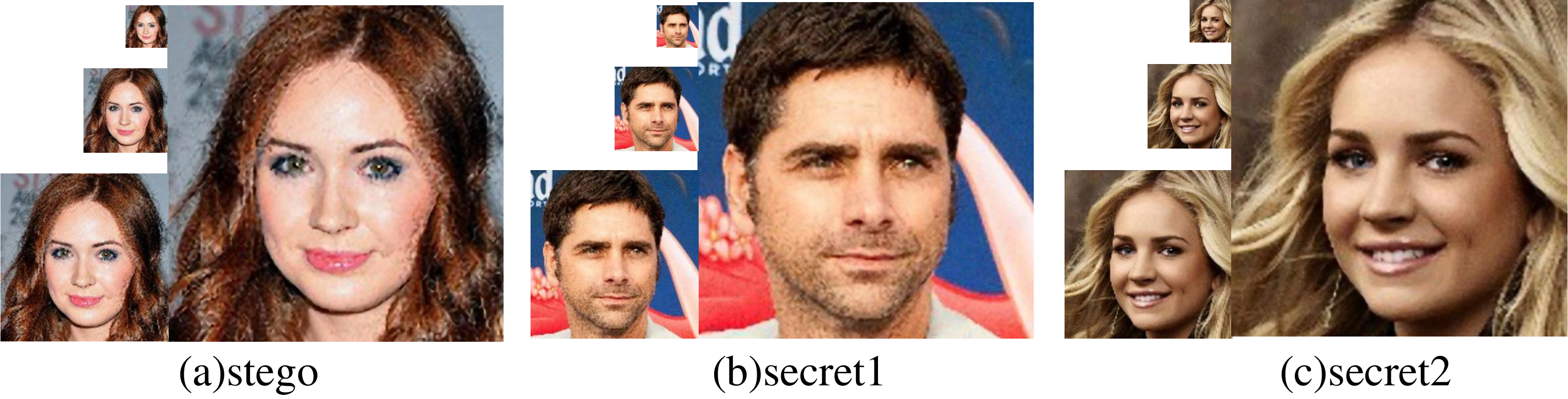}
    \caption{Images sampled at different resolutions.}
    \label{fig8}
\end{figure}

Fig. \ref{fig8} illustrates stego images and two secret images sampled using super-resolution sampling based on a model trained at a resolution of 128$\times$128. The images are shown at resolutions of 64,128, 256, and 512. Due to our utilization of Implicit Neural Representation for image representation, this method exhibits resolution-independent characteristics. This means that image generation and representation are not constrained by specific pixel quantities or resolutions, but rather achieved through encoding and decoding vectors in a high-dimensional latent space. Therefore, we are able to sample stego and secret images at any resolution from models trained at a fixed resolution, showcasing a high level of flexibility and generalization capability in handling images of different resolutions.

\subsection{Robustness}

\begin{figure}[H]
    \centering
    \includegraphics[width=\linewidth]{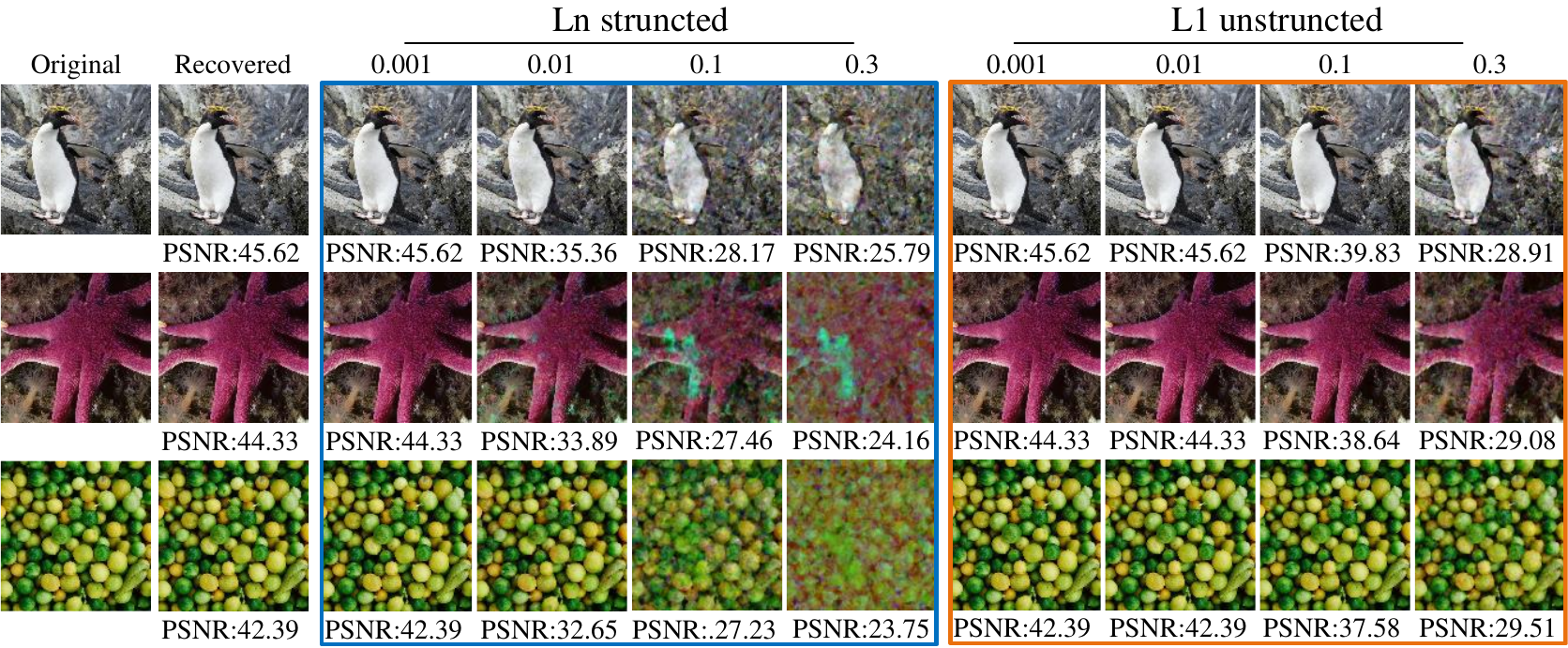}
    \caption{Robustness analysis of stego function.}
    \label{fig9}
\end{figure}

Robustness is another criterion for evaluating image steganography. For robustness analysis, we consider using model pruning to simulate whether the secret image can still be recovered from the steganographic function under malicious attacks. We randomly select images from the Div2k dataset for the experiment, with a resolution of 128$\times$128. We conduct experiments on the stego function using both Ln structured pruning and L1 unstructured pruning. The stego images and secret images recovered before and after pruning are compared to the original images. As shown in Figure \ref{fig9}, we observe the impact of different pruning rates on the recovery quality of stego images and secret images. The first row presents the stego images, and the second and third rows display the two secret images. The results indicate that our stego function has a significantly better resistance to L1 unstructured pruning compared to Ln structured pruning. Moreover, when the pruning rate of the stego function is below 0.1, it adequately presents the stego images and secret images. Although robustness gradually decreases as pruning rates increase, overall, our stego function demonstrates a certain degree of resistance to model pruning.

\subsection{Impact of Weight Selection Ratio on Image Quality}

\begin{figure}[H]
    \centering
    \includegraphics[width=\linewidth]{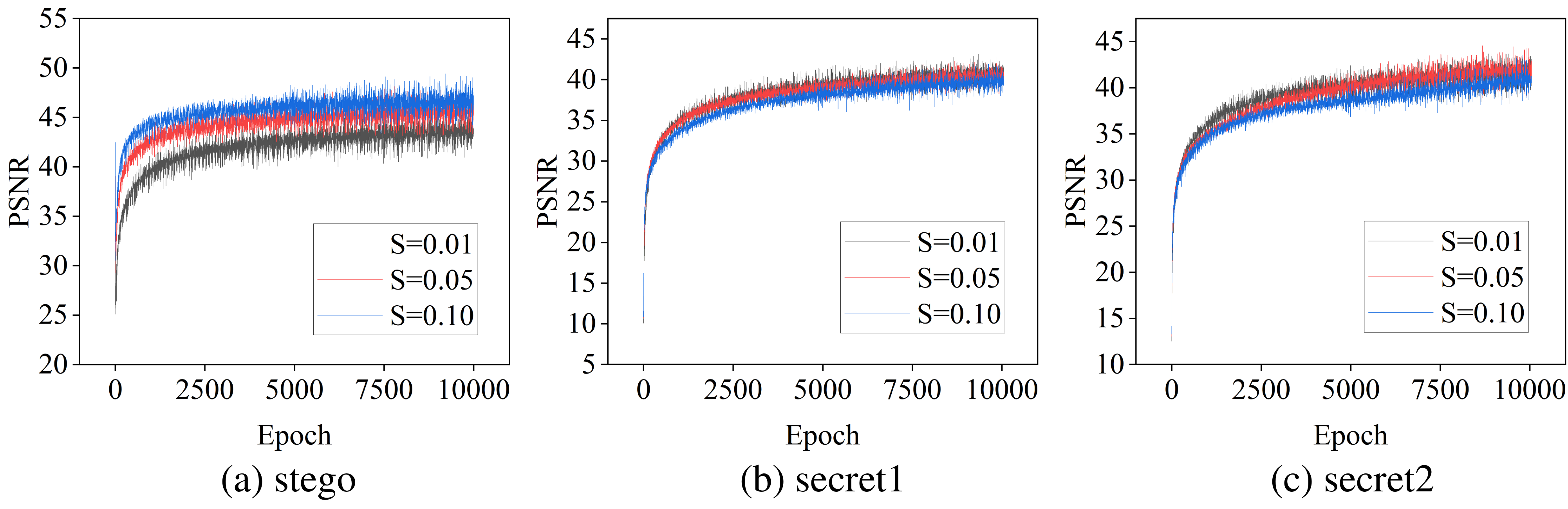}
    \caption{PSNR of stego and secret images under different weight selection ratios.}
    \label{fig10}
\end{figure}

In the experiments conducted, we set the weight selection ratio S to be 0.05. In this section, we varied the weight selection ratio to different values. As shown in Fig. \ref{fig10}, we present the trends in image quality for stego and secret images over the first 10,000 epochs. It can be observed that the quality of stego images is directly proportional to the size of S, while the quality of secret images is inversely proportional to the size of S. This is because as S increases, more important weight parameters for representing stego images are retained, making optimization of stego images easier. However, fewer shared weight parameters are available for optimizing secret images, leading to a decrease in the quality of secret images. To balance the quality of stego and secret images, we chose S as 0.05 as the baseline setting for all experiments.

\subsection{Ablation Study}

\begin{figure}[H]
    \centering
    \includegraphics[width=\linewidth]{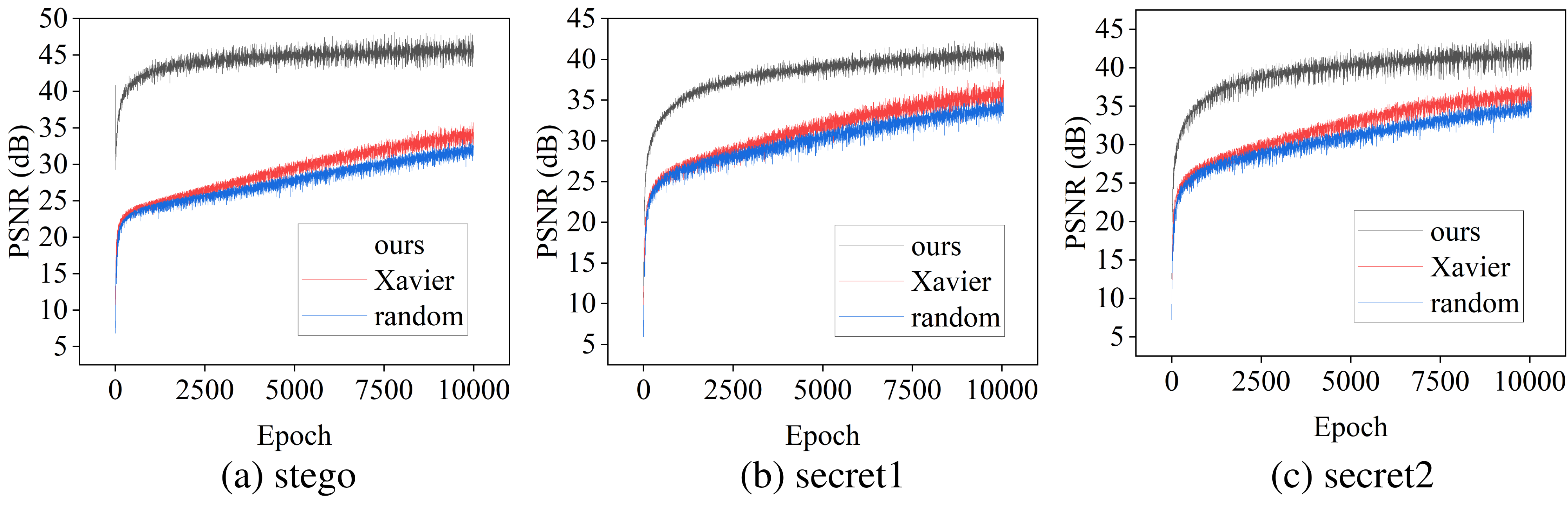}
    \caption{Comparison of PSNR values for three different methods.}
    \label{fig11}
\end{figure}

In order to validate the effectiveness of our pre-training cover images and weight selection method, we designed two ablation contrast experiments without pre-training cover images. The first method involved using Xavier\cite{36} initialization for weight initialization, followed by experiments with the same weight selection strategy. The second method involved no initialization at all, directly selecting weight positions at random for joint training. Fig. \ref{fig11} illustrates the trends in PSNR for stego images (Fig. \ref{fig11}a), the first secret image (Fig. \ref{fig11}b), and the second secret image (Fig. \ref{fig11}c) for the three methods.

It can be observed that our method of pre-training cover images and selecting important weights yielded significantly better quality for both stego and secret images compared to the two contrast methods without pre-training. Specifically, while the Xavier initialization method and random weight position selection method initially showed promising performance, as training progressed, their improvement speed and final results were inferior to our pre-training approach. The significant difference can be attributed to the pre-training process, which enables the cover images to reach a stable and high-quality state, with important weight parameters already adapted to the image features. As a result, in subsequent joint training, these weights can more efficiently optimize stego and secret images. In contrast, without pre-training, weight parameters start learning from scratch, leading to a slower and less stable optimization process, ultimately affecting image quality. In conclusion, our method of pre-training cover images and selecting important weights significantly enhances the quality of stego and secret images, demonstrating the necessity and effectiveness of pre-training and weight selection.

\subsection{Undetectability of Stego Images}

Evaluating the security of stego images through steganalysis is a crucial assessment component in image hiding tasks. Specifically, steganalysis measures the likelihood of distinguishing stego images from cover images using steganalysis tools. We employ two categories of methods to evaluate the undetectability of the stego images generated by our method: traditional statistical methods and novel deep learning-based methods.

(1) Statistical Steganalysis

\begin{figure}[H]
    \centering
    \includegraphics[width=0.8\linewidth]{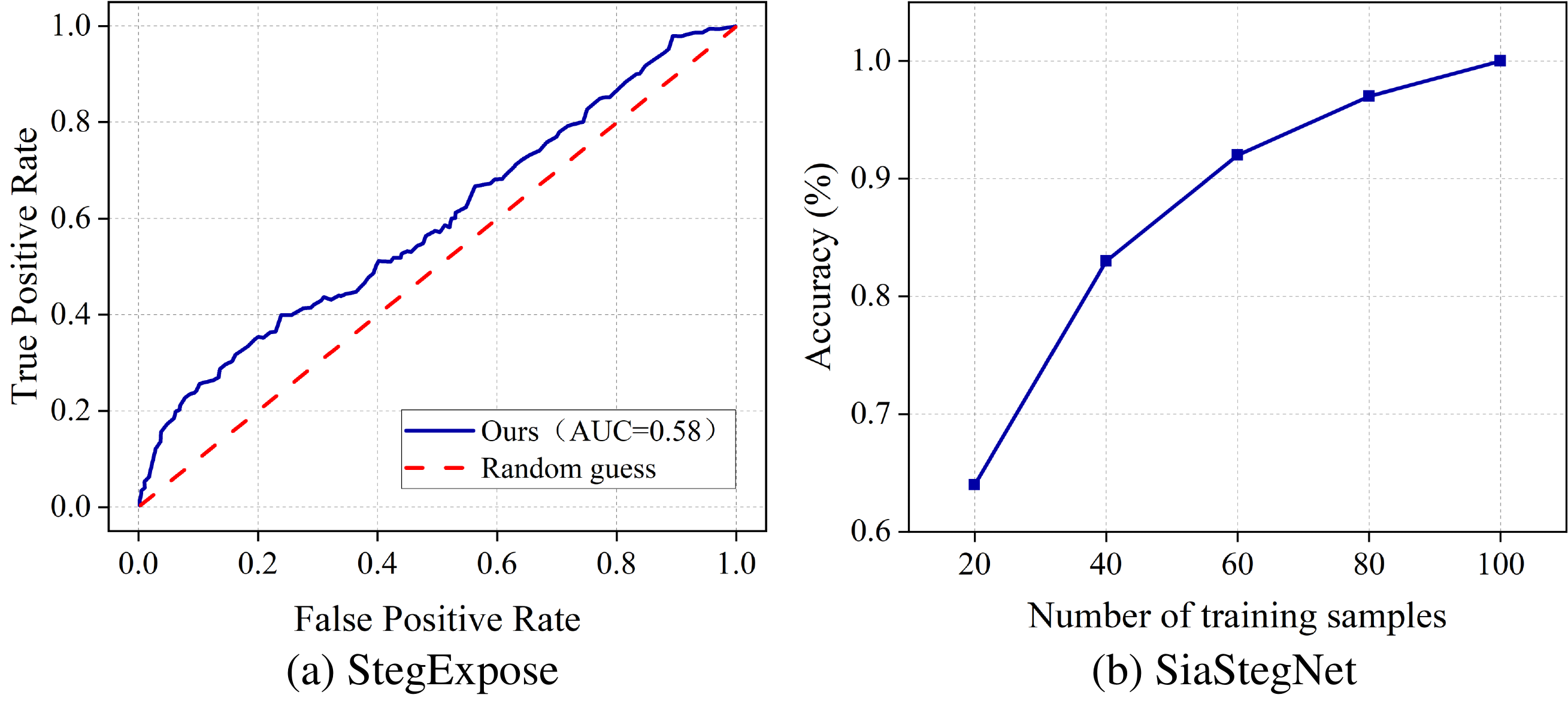}
    \caption{Undetectability of stego images sampled from stego functions for (a) StegExpose and (b) SiaStegNet.}
    \label{fig12}
\end{figure}

We use the open-source steganalysis tool StegExpose41 to measure the anti-steganalysis capability of our model. Following the protocol outlined in reference 5, we randomly select 100 cover images and secret images from the test dataset, generate stego images using our method, and input them into StegExpose for evaluation. By adjusting the detection threshold in StegExpose, we obtain the Receiver Operating Characteristic (ROC) curve. As shown in Fig. \ref{fig12}(a), the Area Under the Curve (AUC) value of our method's ROC curve is 0.58, closer to random guessing (AUC=0.5). This indicates that the stego images generated by our model exhibit high security, with a high probability of deceiving the StegExpose tool.

(2) Deep Learning-Based Steganalysis

SiaSteNet42 is a network used for image steganalysis. Following the protocol in reference 22, we train SiaSteNet with varying numbers of cover/stego image pairs to investigate how many image pairs are needed to enable SiaSteNet to detect stego images. Fig. \ref{fig12}(b) illustrates the detection accuracy of SiaSteNet plotted by changing the number of training image pairs.

It can be observed that in order to accurately detect the presence of secret data, an attacker would need to collect at least 100 labeled cover/stego image pairs, posing a significant challenge in real-world scenarios. This indicates that our method exhibits high resistance to steganalysis, as steganalysis tools are completely unable to differentiate between stego data and cover data. This is because the implicit steganographic alterations do not directly edit the discrete representation of the data, but rather operate on the representation after it has been transformed into a neural network, making it difficult to detect when converted back to a discrete representation.

\subsection{Undetectability of Stego Functions}

Since we aim to covertly embed secret functions into stego functions, it is necessary to conduct analysis to detect the presence of secret functions within stego functions transmitted through public channels. Unfortunately, almost all existing steganalysis tools are designed for steganalysis of multimedia data (images/videos/text). We adopt three steganalysis strategies specifically targeting stego functions based on our experience.

(1) Weight Distribution

\begin{figure}[H]
    \centering
    \includegraphics[width=\linewidth]{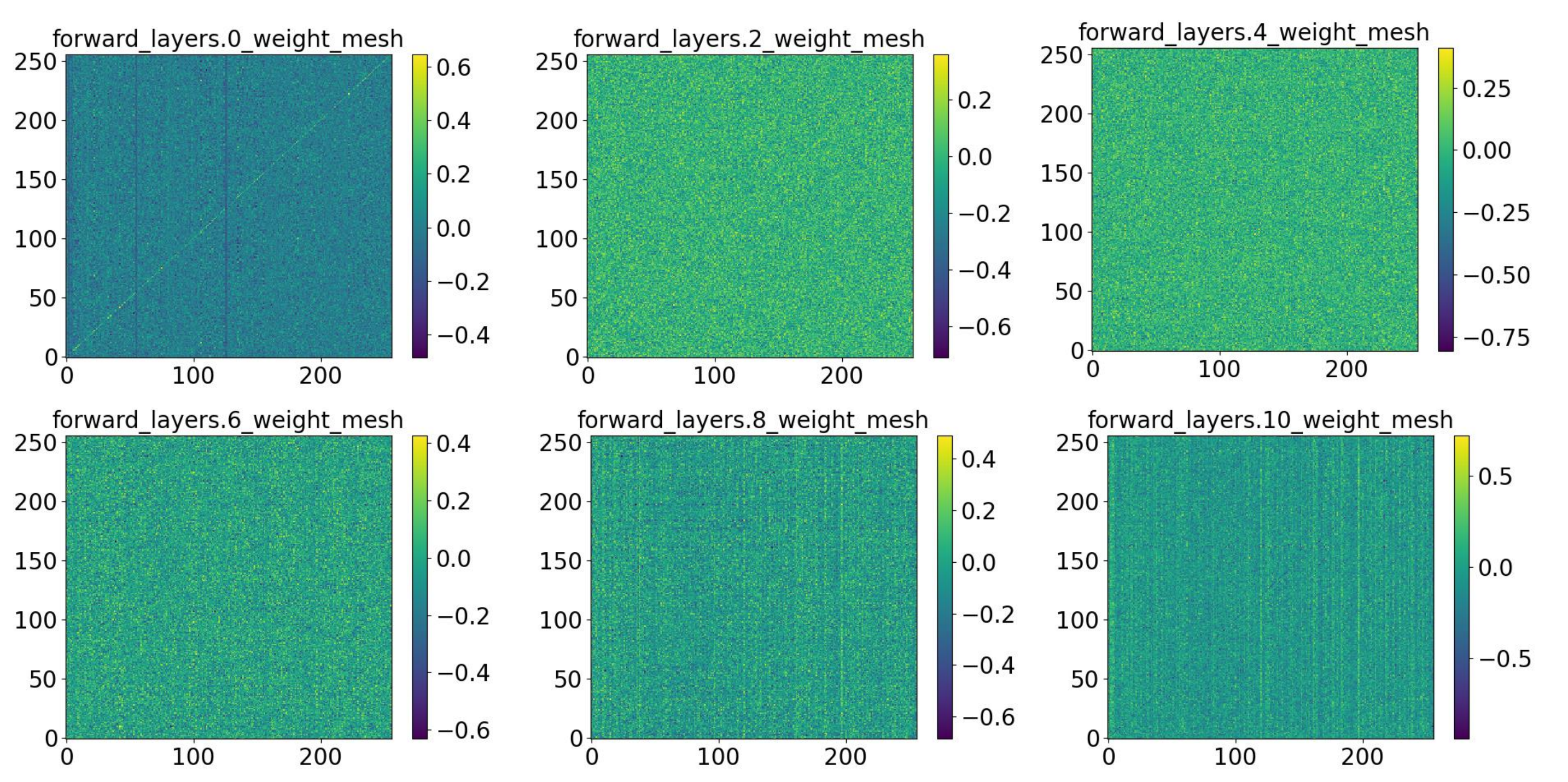}
    \caption{Visualization of the weight distribution of stego function.}
    \label{fig13}
\end{figure}

We assess the regularity of weight distribution and the concealment of steganography by observing the weight spacings of stego data. Fig. \ref{fig13} illustrates the weight distribution at different layers of the stego functions, where most weights exhibit a relatively uniform distribution, effectively resisting steganalysis attacks based on statistical methods.

(2) SVM Classification Detection

We establish a model pool consisting of 100 stego functions and 100 clean functions with the same structure to train a classifier to detect differences in parameter distribution between stego functions and clean functions. Each function is trained on 100 different images, and stego functions and clean functions are trained on carrier images with the same implicit representation. We calculate a 100-bin histogram for the parameters of each function to obtain 100-dimensional feature vectors for training Support Vector Machine (SVM). We randomly select 80 pairs of features from stego functions and clean functions for training, while the remaining 20 pairs are used for testing. The detection accuracy, as shown in Table \ref{tab3}, indicates that the existence of secret functions cannot be detected from stego functions, with detection accuracy close to 50\%. Here, "SVM-poly" refers to SVM with a polynomial kernel, while "SVM-rbf" refers to SVM with a Gaussian (RBF) kernel.

\begin{table}[H]
    \centering
    \caption{The undetectability of the proposed method.}
    \begin{tabular}{cc}
    \hline
         Detection method&Detection rate\\
         \hline
SVM-poly&0.53\\
SVM-rbf&0.54\\
\hline
    \end{tabular}
    \label{tab3}
\end{table}

3) Secret Function Leakage Analysis

\begin{figure}[H]
    \centering
    \includegraphics[width=0.8\linewidth]{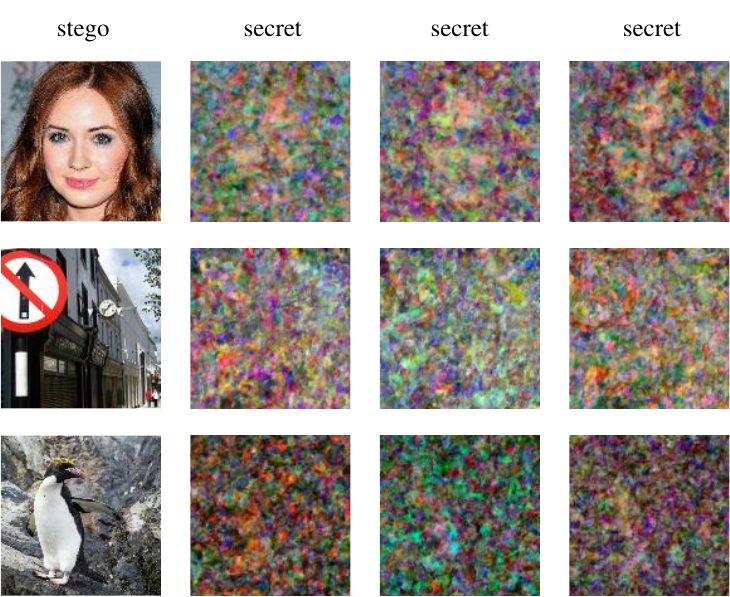}
    \caption{Visualization of Secret Images Stolen using a Random Key Attack.}
    \label{fig14}
\end{figure}

Our approach allows a recipient with the correct key to extract the secret function and recover the secret image from the publicly available stego functions. However, for attackers, they may attempt to crack the secret function by using randomly guessed keys to attack the stego functions. To test the resistance of stego functions to such attacks, we envision a scenario where attackers have knowledge of the fixed weight positions (i.e., edge information) and attempt to replace secret weights using 1000 different random seeds to recover the secret image. This method simulates the potential random guessing attack strategy that attackers may employ. Fig. \ref{fig14} presents the results of three random attack experiments conducted on three different datasets, with the first column showing the stego image and the subsequent three columns displaying the secret images obtained from the three random attacks. It can be clearly seen that even when the positions of the weights are known, attackers are almost unable to extract any meaningful secret images from the stego function using random keys. The images generated by each attack exhibit high randomness and unrecognizable patterns, demonstrating the effective protection of the security of the secret information. This indicates that attackers face significant challenges in launching successful attacks by attempting to extract secret images from stego functions by triggering secret function extraction using random keys.

\subsection{Steganographic Scheme Comparison}

To validate the superiority of our proposed implicit representation steganographic method, our approach is compared with traditional image hiding methods such as LSB\cite{43}, as well as deep learning-based steganographic methods including HiDDeN\cite{7}, Baluja\cite{19}, Weng et al\cite{23}, MISDNN\cite{20}, ISN\cite{21} and DeepMIH\cite{22}. Additionally, we compare our method with a similar implicit representation steganographic scheme, StegaINR\cite{16}. In order to facilitate a fair comparison among different types of information hiding schemes, we replicate the results of LSB, HiDDeN, Baluja, Weng et al., MISDNN and ISN as tested on 256$\times$256 resolution images from the COCO and Div2k dataset\cite{38} as reported in the DeepMIH literature. Subsequently, both our method and StegaINR are retrained using images from the COCO and Div2k dataset at a resolution of 256$\times$256. The secret function structure of StegaINR comprises three hidden layers with 64 neurons each, while the stego function consists of three hidden layers with 256 neurons each. The experimental comparison results are presented in Table \ref{tab4}.

\begin{table}[htbp]
    \centering
    \caption{Comparison of image quality among different steganographic schemes.}
    \belowrulesep=0pt
    \aboverulesep=0pt
    \begin{tabular}{c|cccc|cccc}
\hline \multirow{3}{*}{Methods} & \multicolumn{8}{c}{Cover/Stego-1 image pair} \\
\cmidrule{2-9} & \multicolumn{4}{c}{COCO} & \multicolumn{4}{|c}{Div2k} \\
\cmidrule{2-9} & PSNR  $\uparrow$  & SSIM  $\uparrow$  & RMSE  $\downarrow$  & MAE  $\downarrow$  & PSNR  $\uparrow$  & SSIM  $\uparrow$  & RMSE  $\downarrow $ & MAE  $\downarrow$  \\
\hline LSB & 34.82 & 0.9561 & 6.55 & 5.36 & 35.03 & 0.9611 & 5.30 & 6.49 \\
 HiDDeN & 37.24 & 0.9791 & 8.11 & 5.94 & 37.52 & 0.9712 & 5.90 & 8.41 \\
Baluja & 36.99 & 0.9575 & 7.72 & 5.83 & 37.25 & 0.9587 & 7.65 & 5.73 \\
 Weng et al. & 36.73 & 0.9672 & 5.78 & 4.06 & 39.57 & 0.9788 & 3.08 & 4.25 \\
 DeepMIH & 40.30 & 0.9805 & 4.14 & 2.83 & 43.72 & 0.9895 & 2.81 & 1.94 \\
 StegalNR & 16.97 & 0.3046 & 9.65 & 16.91 & 19.22 & 0.4823 & 8.95 & 15.26 \\
 ours & 41.26 & 0.9921 & 2.65 & 1.91 & 40.94 & 0.9903 & 2.87 & 1.08 \\
\hline & \multicolumn{8}{c}{Secret/Recovery-1 image pair} \\
\cmidrule{2-9} \multirow[t]{2}{*}{Methods} & \multicolumn{4}{c}{COCO} & \multicolumn{4}{|c}{Div2k} \\
\cmidrule{2-9} & PSNR  $\uparrow$  & SSIM  $\uparrow$  & RMSE  $\downarrow$  &  MAE $\downarrow $ & PSNR $ \uparrow$  & SSIM  $\uparrow$  & RMSE  $\downarrow$  &  MAE $\downarrow$  \\
\hline LSB & 24.96 & 0.8939 & 17.93 & 15.31 & 24.99 & 0.8951 & 18.16 & 15.57 \\
 HiDDeN & 33.70 & 0.9443 & 6.13 & 4.55 & 36.10 & 0.9648 & 5.80 & 4.36 \\
 Baluja & 34.85 & 0.9558 & 7.34 & 5.33 & 36.34 & 0.9716 & 6.01 & 4.47 \\
 Weng et al. & 35.13 & 0.9475 & 6.70 & 4.66 & 37.43 & 0.9487 & 4.96 & 3.58 \\
 MISDNN & 28.21 & 0.8260 & 16.55 & 12.12 & 28.17 & 0.8301 & 16.87 & 12.59 \\
 ISN & 36.09 & 0.9510 & 6.48 & 5.09 & 37.21 & 0.9627 & 5.55 & 4.01 \\
 DeepMIH & 36.55 & 0.9613 & 6.04 & 4.09 & 41.22 & 0.9838 & 3.83 & 2.58 \\
 StegaINR & 27.32 & 0.9026 & 3.72 & 9.02 & 30.56 & 0.9201 & 3.01 & 8.29 \\
 ours & 38.24 & 0.9885 & 2.98 & 1.21 & 37.65 & 0.9801 & 2.99 & 1.35 \\
\hline & \multicolumn{8}{c}{Cover/Stego-2 image pair} \\
\cmidrule{2-9} \multirow[t]{2}{*}{Methods} & \multicolumn{4}{c}{COCO} & \multicolumn{4}{|c}{Div2k} \\
\cmidrule{2-9} & PSNR  $\uparrow$  & SSIM  $\uparrow$  & RMSE  $\downarrow$  &  MAE$\downarrow$  & PSNR  $\uparrow$  & SSIM  $\uparrow$  & RMSE  $\downarrow$  &  MAE $\downarrow$  \\
\hline LSB & 34.77 & 0.9550 & 6.60 & 5.40 & 34.89 & 0.9597 & 6.58 & 5.38 \\
 HiDDeN & 34.19 & 0.9420 & 9.23 & 6.99 & 34.75 & 0.9439 & 9.38 & 6.87 \\
 Baluja & 36.05 & 0.9478 & 8.52 & 6.48 & 36.35 & 0.9493 & 8.30 & 6.26 \\
 Weng et al. & 34.11 & 0.9467 & 7.51 & 5.31 & 36.51 & 0.9627 & 5.76 & 4.20 \\
 MISDNN & 33.01 & 0.8923 & 9.83 & 7.52 & 33.95 & 0.9019 & 8.99 & 6.99 \\
 ISN & 36.98 & 0.9598 & 7.92 & 5.75 & 40.19 & 0.9867 & 5.98 & 4.25 \\
 DeepMIH & 37.21 & 0.9624 & 5.90 & 3.95 & 41.22 & 0.9838 & 3.83 & 2.58 \\
 ours & 38.09 & 0.9822 & 3.17 & 2.37 & 38.25 & 0.9830 & 3.01 & 2.62 \\
\hline & \multicolumn{8}{c}{Secret/Recovery-2 image pair} \\
\cmidrule{2-9} \multirow[t]{2}{*}{Methods} & \multicolumn{4}{c}{COCO} & \multicolumn{4}{|c}{Div2k} \\
\cmidrule{2-9} & PSNR  $\uparrow$  & SSIM  $\uparrow$  & RMSE  $\downarrow$  &  MAE$\downarrow$  & PSNR  $\uparrow$  & SSIM  $\uparrow$  & RMSE  $\downarrow$  &  MAE$\downarrow$  \\
\hline LSB & 13.16 & 0.5311 & 68.79 & 58.90 & 13.04 & 0.5568 & 69.38 & 60.29 \\
 HiDDeN & 33.87 & 0.9469 & 5.98 & 4.47 & 37.02 & 0.9659 & 4.88 & 3.82 \\
 Baluja & 34.95 & 0.9670 & 7.32 & 5.31 & 36.62 & 0.9743 & 5.58 & 4.18 \\
 Weng et al. & 35.82 & 0.9567 & 6.19 & 4.33 & 37.87 & 0.9578 & 4.59 & 3.34 \\
 MISDNN & 29.43 & 0.8708 & 13.99 & 10.43 & 29.45 & 0.8679 & 14.02 & 10.63 \\
 ISN & 36.86 & 0.9525 & 6.24 & 4.88 & 37.81 & 0.9625 & 4.97 & 3.57 \\
 DeepMIH & 37.72 & 0.9696 & 5.53 & 3.72 & 42.53 & 0.9858 & 2.93 & 1.98 \\
 ours & 36.32 & 0.9788 & 3.89 & 2.71 & 37.57 & 0.9798 & 3.52 & 2.56 \\
\hline
\end{tabular}
    
    \label{tab4}
\end{table}

Results show that our method achieves high levels of quality in both hidden and stego images. Furthermore, our approach only requires parameterization of individual data samples without the need for training on large datasets. By leveraging implicit neural representations for steganography, the Implicit Neural Representation (INR) allows for encoding or processing data through a continuous and differentiable function via neural network parameters. This enables the presentation of data at various resolutions using a single neural network. Training data with an implicit representation not only eliminates direct dependence on data resolution but also improves training efficiency while saving memory.

\section{Conclusion}

The proposed steganography by implicit neural representation for multi-image hiding  (StegaINR4MIH) framework, which pre-trains the implicit neural representation function of cover images and utilizes magnitude-based weight selection and secret weight replacement strategies, successfully achieves efficient steganography and independent extraction of multiple images through a single neural network function. This framework not only optimizes the quality of stego images but also ensures the security and concealment of secret images. Moreover, our method enhances the security of secret information effectively through different key triggering mechanisms, ensuring effective protection of secret information even during transmission through public channels. Experimental results demonstrate the favorable performance of our method in distortion assessment, capacity, and security. Implicit neural representation steganography, as a novel research direction, will be further developed in future work to enhance the robustness of the framework by more effectively refining the network structure of INR.

\subsection* {Code, Data, and Materials Availability} 

\begin{itemize}
    \item The archived version of the code described in this manuscript can be freely accessed through GitHub [https://github.com/Haifeng-Jiang/StegaINR4MIH].
\end{itemize}

\subsection* {Acknowledgments}
This work is supported by the General Program of the National Natural Science Foundation of China (62272478), the National Natural Science Foundation of China (61872384,62102451) and Science and Technology Innovation Team Innovative Research Project (Grant No. ZZKY2022102). We have no relevant financial interests in the manuscript and no other potential conflicts of interest to disclose.

\bibliographystyle{spiejour}   
\bibliography{report}   


\vspace{2ex}\noindent\textbf{Weina Dong} is currently an MS candidate at Engineering University of PAP. Her current research interests include information hiding and image steganography.

\textbf{Jia Liu} received the M.S. degree in cryptography from the Engineering University of PAP, Xi’an, China, in 2007, and the Ph.D. degree in pattern recognition and intelligent system from Shanghai Jiao Tong University, Shanghai, China, in 2012. He is currently an Associate Professor with the Key Laboratory of Network and Information Security, Engineering University of PAP. His research interests include machine learning and information security.

\listoffigures
\listoftables

\end{spacing}
\end{document}